\documentclass[letterpaper, 10 pt, conference]{ieeeconf}
\IEEEoverridecommandlockouts                              

\pdfoutput=1    

\usepackage{cite}
\usepackage{amsmath,amssymb,amsfonts}
\usepackage{algorithmic}
\usepackage{subcaption}
\usepackage{graphicx}
\usepackage{textcomp}
\usepackage{xcolor}
\usepackage{float}
\usepackage{amsmath}
\usepackage{hyperref}

\def\BibTeX{{\rm B\kern-.05em{\sc i\kern-.025em b}\kern-.08em
    T\kern-.1667em\lower.7ex\hbox{E}\kern-.125emX}}
\begin{document}

\title{\LARGE \bf
Quantification of Actual Road User Behavior\\on the Basis of Given Traffic Rules
}

\author{Daniel Bogdoll$^{1,*}$, Moritz Nekolla$^{1,*}$, Tim Joseph$^{1}$, J. Marius Zöllner$^{1}$
\thanks{$^{1}$FZI Research Center for Information Technology, Karlsruhe, Germany
        {\tt\small bogdoll@fzi.de}}%
}



\maketitle
\begingroup\renewcommand\thefootnote{\textasteriskcentered}
\footnotetext{These authors contributed equally}
\endgroup

\maketitle

\begin{abstract}
Driving on roads is restricted by various traffic rules, aiming to ensure safety for all traffic participants. However, human road users usually do not adhere to these rules strictly, resulting in varying degrees of rule conformity. Such deviations from given rules are key components of today's road traffic. In autonomous driving, robotic agents can disturb traffic flow, when rule deviations are not taken into account. In this paper, we present an approach to derive the distribution of degrees of rule conformity from human driving data. We demonstrate our method with the Waymo Open Motion dataset and \textit{Safety Distance} and \textit{Speed Limit} rules.

\end{abstract}


\section{Introduction}
\label{sec:introduction}
Human traffic participants have the ability to slightly bend traffic rules in order to keep traffic flowing or to react to special situations. While strong violations of rules are unacceptable, slight twists regarding the rule conformity (RC) are socially accepted. Especially in corner case situations~\cite{Bogdoll_Description_2021_ICCV}, a flexibility that goes beyond enforced traffic rules is often necessary. Unfortunately, this knowledge about human behavior is not freely available but can be extracted from datasets. In informed machine learning~\cite{rueden_informed_2021, kiw_sota} such knowledge can be of great value. While Deep Learning brought unmatched performance, many questions such as rule enforcement, explainability or plausibility checks remain open, driving the field of informed machine learning. For example, an autonomous vehicle system can be enriched with knowledge about human behavior to improve the overall traffic flow. However, prior knowledge needs to be picked carefully. If it narrows down the landscape of solutions too much, the benefits of deep learning diminish. \\

While there are traffic rules that offer a lot of room for interpretation and can only be used as a source of knowledge in a more complex way, many traffic rules are directly quantifiable and therefore a suitable knowledge source. However, human drivers do not strictly follow these rules in the real world. Their behavior differs from situation to situation, since they utilize their prior experience. That process of valuing the importance of each rule at any given time enables humans to properly deal with a wide variety of situations. Therefore, static traffic rules can be formally enriched based on human interpretation. Collin et al.~\cite{collin_safety_rulebooks} provide a framework based on \cite{rulebooks} for rule-based planning which incorporates rule deviations. However, this allowed “degree of violation […] with respect to the rule” is not described and left to the user to set. This open research question is a core motivation for our work.

\begin{figure}[tp]
\includegraphics[scale=0.295]{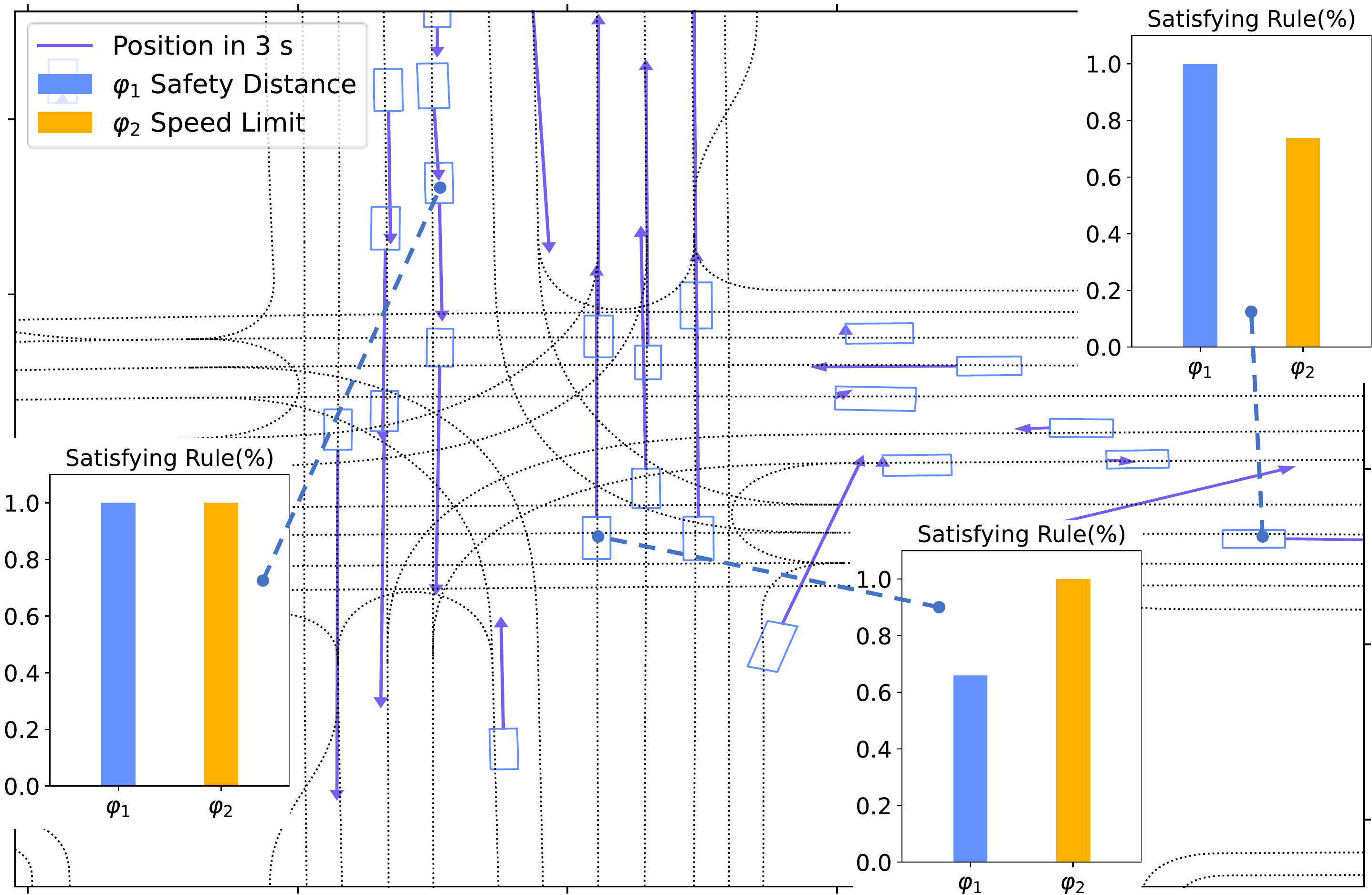} 
\caption{Scenario 1, scene 71 in the \textit{validation.tfrecord-00018-of-00150} file from the Waymo Open Motion Dataset shows traffic at an intersection. $\varphi_1$ represents the \textit{Safety Distance} rule and $\varphi_2$ refers to \textit{Speed Limit} rule. The histograms show the degrees of rule conformity for selected vehicles.}
\label{fig:slack_dist}
\end{figure}

\textbf{Research Gap.} Existing research most often extracts knowledge from data without being able to examine certain aspects of it \cite{nvidia_e2e,Kuderer2015LearningDS}. The only closely related work from Cho and Oh~\cite{cho_learning-based_2018} that addresses the specific topic of degrees of rule conformity does only perform the quantification based on an in-house simulation that does not provide data about actual human behavior. Additionally, they neither disclose their algorithms nor the results.

\textbf{Contribution.} Quantifying knowledge about human behavior in respect to traffic rules is the core contribution of this paper. For this purpose, we demonstrate our method by analyzing the expert behavior from the Waymo Open Motion dataset~\cite{sun2020scalability} based on the rules \textit{Safety Distance} and \textit{Speed Limits}. The code is available at \url{https://url.fzi.de/iv2022}

In Chapter \ref{sec:related_work}, we give an overview over previous work. In Chapter \ref{sec:method}, we describe the traffic rules and how we extracted human behavior distributions regarding these rules from the dataset. Finally, in chapter \ref{sec:evaluation} we discuss the results. 


\section{Related Work}
\label{sec:related_work}

\begin{figure*}[tp]
    \includegraphics[scale=0.65]{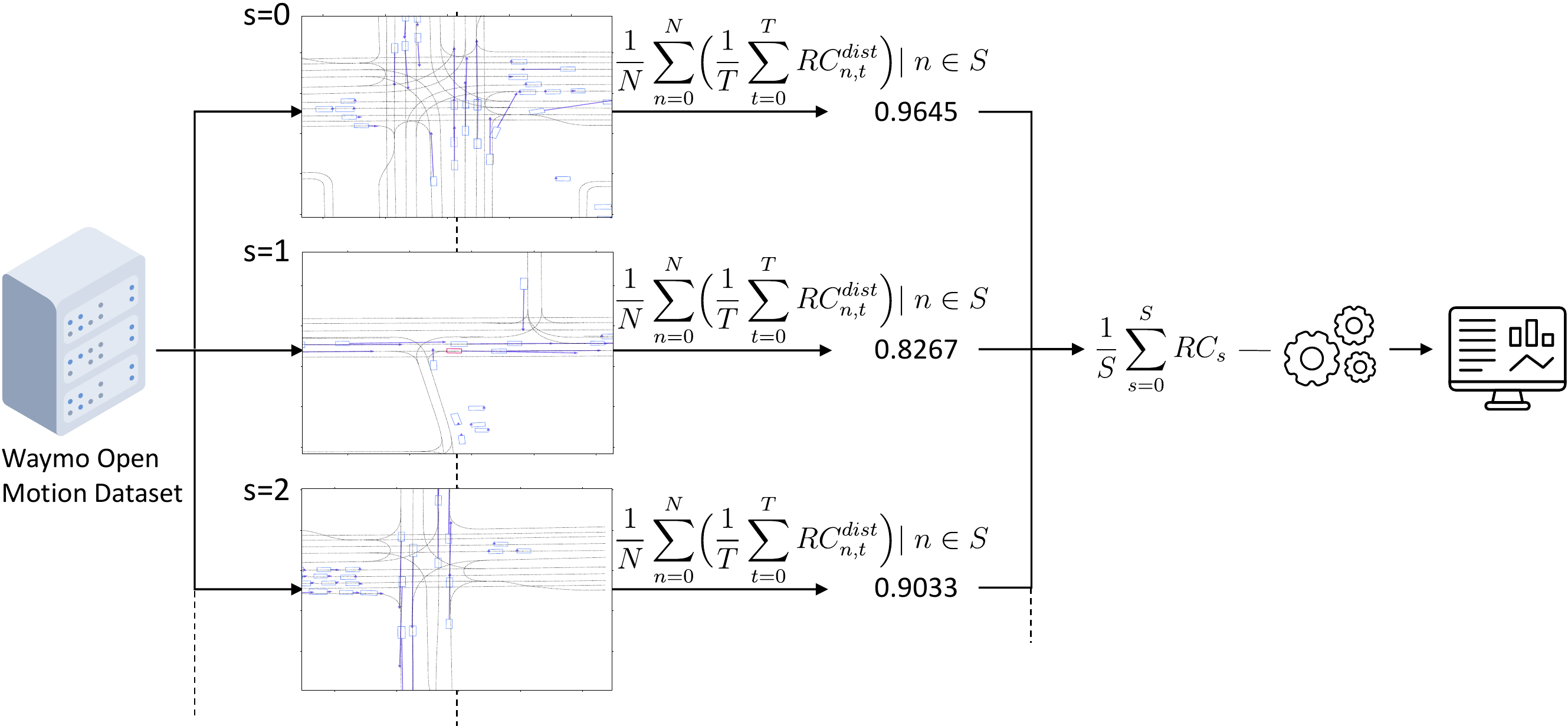} 
    \caption{Overview of the quantification process for the \textit{Safety Distance} rule. Within every scenario $s$ of the dataset, which consist of $T$ frames each, we first calculate the score $RC^{dist}_{n,t}$ per vehicle $n$ and frame $t$. Then, we compute the average score $RC^{dist}_{n,s}$ for each vehicle $n$ over all frames. Now, we are able to compute the average over all vehicles $RC^{dist}_{s}$ as the average of a whole scenario. Finally, this allows us to compute the average score for the whole dataset $RC^{dist}_{total}$. The intermediate steps allow us to describe single vehicles, sceneOverviews, scenarios and to generate the behavioral distribution for the whole dataset.}
    \label{fig:blueprint}
\end{figure*}

 In autonomous driving, datasets are in many ways the backbone of current developments, be it for their usage as training data or their ability to benchmark algorithms and models. Since the release of the KITTI dataset~\cite{Geiger2012CVPR} in 2012, many more datasets with increased size and variety have emerged~\cite{addatasets}. The process of analyzing datasets to gain further insights has a wide range of applications, often with the aim to extract knowledge from them. To develop autonomous vehicles that show more human-like behavior, which includes enhanced flexibility in many types of situations, datasets can be utilized to extract the observed human behavior in the data. \\
 
 \textbf{Abstract Knowledge Extraction.} There exist many ways to utilize knowledge from datasets for the planning task of an autonomous vehicle. Supervised imitation learning can utilize recordings as training data for Convolutional Neural Networks (CNN) to learn to directly predict steering commands from sensor data of the ego vehicle~\cite{nvidia_e2e}. In the related field of Inverse Reinforcement Learning~\cite{Kuderer2015LearningDS}, existing data is used to learn the reward function for a Reinforcement Learning agent based on the observed behavior of road users. Prediction methods~\cite{Ye2021TPCNTP} are among the most prominent methods in this field, as their core task is to predict human behavior. Predictions can also be applied to the ego-vehicle as a reference trajectory to follow or as an additional input for the planning module~\cite{aurora_cvpr}, since own actions and the actions of other road users are intertwined. In~\cite{rezaee2021drive}, constraints were extracted from human driving trajectories. These constraints can be combined with simple cost functions in motion planning, enabling interpretability and more human-like behavior. In contrast to these approaches, which only extract abstract behavior, our method is able to quantify and assign knowledge to general concepts, such as traffic rules. \\
 
 
\textbf{Traffic Rule Violations.} The intent of violating traffic rules was first analyzed in~\cite{violate_emilio} based on questionnaires, not representing actual behavior from recorded data. In~\cite{violate_china}, data from the “Automated Enforcement System” in China was analyzed. This system can detect 12 different traffic rule violation types. The authors focus on the location-based type and frequency of such violations. Similarly, in~\cite{helou2021reasonable}, the frequency of violations in a simulation was analyzed. These methods focus on geographical distributions of traffic rule violations. While similar to the aforementioned methods in the sense that they deal with traffic rule violations, our approach is directly linked to the behavior of individuals and the scenarios they act in. This allows for an analysis of the severity of violations related to drivers. \\

\textbf{Rule-based Learning.} In~\cite{Puranic2020LearningFD,learn_demo}, the idea of rule-based learning from demonstrations was investigated. The authors rated expert demonstrations based on given rules, formulated with the signal temporal logic language. This way, it was possible to only learn primarily from demonstrations that fulfil a defined quality. They demonstrated their approach with small, self-designed experiments instead of applying it to real-world data. 
The idea of “data driven rule books” was explored in \cite{Censi_Slutskyy_Devi_Xuan_Chen_2021}, where “a model may be generated based on human driving behavior that governs how an autonomous vehicle maneuvers in that scenario”. Unfortunately, there is no scientific work based on the patent that would reveal more details about the procedure. The idea of deriving values for rule conformity from given traffic rules was originally developed in \cite{cho_learning-based_2018} and continued in \cite{cho_deep_2019} in order to overcome dilemma situations and generate more human-like behavior. The authors formulated traffic rules as signal temporal logic (STL) formulas, which enables the generation of satisfaction margins for any given rule. This margin or “robustness slackness” is calculated by observing expert behavior. Their quantification is based on an in-house simulation that does not provide data about actual human behavior. Beyond their conceptual approach, our work enables behavioral analysis of human drivers based on real data and discusses the distribution of driving behaviors in the Waymo Open Motion Datasets.


\section{Method}
\label{sec:method}
The core of our work is the ability to derive the distribution of human behavior in relation to given rules from real world datasets. For this quantification, we distinguish between global and local rule conformity. While local rule conformity is dependent on a specific situation, that lead to the deviation of a rule, the global degree of conformity is the distribution over a whole scenario or dataset. In this work, we propose a method for the quantification of global rule conformity. Rule conformity describes the degree of compliance with a traffic rule.

In our work, our aim is to provide global rule conformity distributions and average values $RC^{}_{}$ on scenario- and dataset levels. These describe the compliance with which the drivers have followed traffic rules. In this paper, we exemplarily calculated the scores for the Waymo Open Motion Dataset~\cite{sun2020scalability} as shown in Figure~\ref{fig:blueprint}. It consists of 103,354 segments, each containing 20 seconds of footage from urban to suburban terrain from six different cities among the USA. The segments were further broken down into scenarios of different lengths. We follow the definitions of scene and scenario by Ulbrich~et~al. \cite{Ulbrich:2015d}, where a scene is a single frame $t$, and a scenario $s$ consists of successive scenes.

The Waymo dataset consists of detailed object lists with corresponding map data. While we are generally also interested in behavior on German roads, there is no large-scale motion data set with corresponding map data that has been recorded in Germany. While we are aware of motion datasets (partly) recorded in Germany, such as \cite{interactiondataset,inDdataset,Fleck:2018}, these are insufficient in size and variety for our quantification approach. However, since the traffic laws in the USA are similar to those in Germany, we deem the Waymo dataset appropriate, keeping the domain gap small. For performance reasons, we analyze the dataset at a rate of 1~Hz instead of the provided 10~Hz. Since our code is open-source, all parameters can be adjusted to rerun the analysis for enhanced precision.


\subsection{Rule Conformity for Safety Distances}
\begin{figure}[h]
\centering
\includegraphics[width=0.49\textwidth]{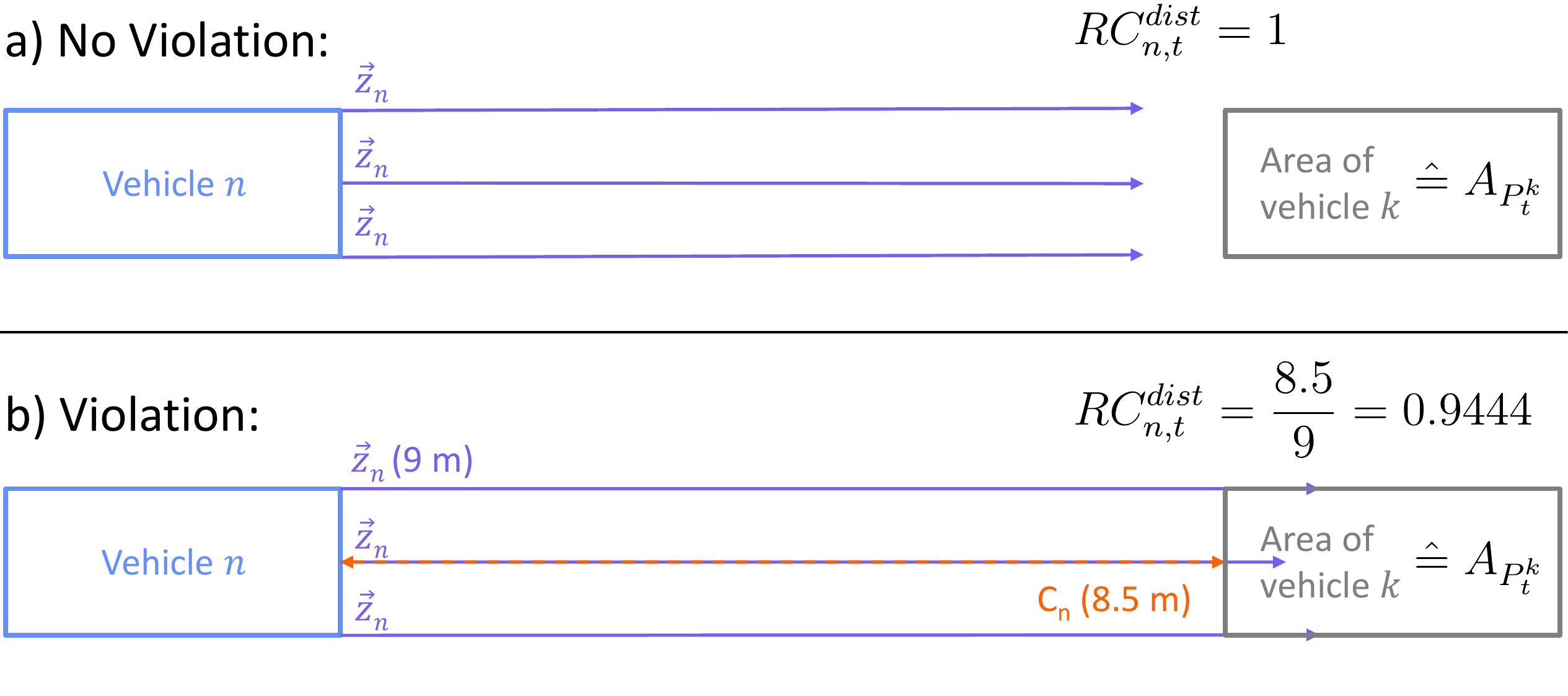}    
\caption{Computation of the \textit{Safety Distance} rule conformity degree. In a), no rule violation occurs. In b), a violation is shown, where  $|\vec z_n| = 9~m$ is the distance according to the three-second-rule and $c_n=8.5~m$ the distance between the ego vehicle and the intersection point between $\vec z_n$ and the vehicle ahead.}
\label{fig:vis_safety_distance}
\end{figure}

A \textit{Safety Distance} between vehicles in Germany is regulated in the road traffic regulations (Straßenverkehrs-Ordnung (StVO)), §4~cl.~1: “As a rule, the distance to a vehicle in front must be large enough to allow stopping behind it even if it brakes suddenly.”~\cite{stvo}. To fulfil this requirement, we implemented the three-second rule~\cite{karlsson_encoding_2021}, which is a recommended practice in the USA, thus bridging the domain gap between German laws and data recorded in the USA. Mathematically expressed, in a scenario $s$ at time $t$, for each  driver $n$, their position $P^{n}_t$ is multiplied by three times their velocity vector $\vec v_{n,t}$ to receive the future position $P^{n}_{t+1}$. $P$ denotes the center foremost point of a vehicle. That forms the vector $\vec z_{n,t} = (P^{n}_t,P^{n}_{t+1})$ which expresses the distance the vehicle would travel in one direction within three seconds at constant speed. A violation occurs if there exists an intersection between the area $A_{P^{k}_t}$ covered by another vehicle $k$ with position $P^{k}_t$ and $\vec z_{n}$. Then, $c_{n,k,t}$ denotes the distance between $P^{n}_t$ and the intersection point of $\vec z_{n}$ and $A_{P^{k}_t}$. In case of multiple intersections, the closest vehicle is taken into account. As shown in Figure~\ref{fig:vis_safety_distance}, we calculate the rule conformity degree $RC^{dist}_{n,t}$ for driver $n$ as follows:

\begin{equation}
RC^{dist}_{n,t}=\begin{cases}
\begin{aligned} 
  &\frac{c_{n,k,t}}{|\vec z_{n,t}|} \quad \text{if } \,\exists k: A_{P^{k}_t}\,\cap \,\vec z_{n,t} \mid k \ne n\\
  &1 \hspace{2.5em} \text{else}
\end{aligned}
\end{cases}
\end{equation}

The equation ensures that $RC^{dist} \in [0,1]$ at all times. Besides the center point of each car, this procedure is repeated with the left and right edges as well, as shown in Figure~\ref{fig:vis_safety_distance}. That is, to take into account curves and cars that are slightly offset from each other.

Accumulating these individual values, we retrieve an average for every scenario $s$ in the dataset including $N$ drivers in $T$ frames. A scenario itself consists of a road graph with several frames that contains drivers and their properties:
\begin{equation}
    RC^{dist}_{s} = \frac{1}{N}\sum_{n=0}^{N}\Bigl(\frac{1}{T}\sum_{t=0}^{T}RC^{dist}_{n,t}\Bigl)
\end{equation}
For the scene shown in Figure~\ref{fig:slack_dist}, $RC^{dist}_{s}\approx0.9098$. In other words, the drivers bend the \textit{Safety Distance} rule, on average, by around 10\%.

We apply several pre-processing steps on the data to avoid a distortion of the resulting distribution. Vehicles below a velocity of $5~km/h$ are discarded from the calculations, since they are typically either stuck in a traffic jam, waiting at a red light or parking. Since the projected position for the calculation of the \textit{Safety Distance} does not incorporate lanes, only vehicles that are heading in a similar direction are taken into account, allowing an angular deviation of $20~\%$. Figure \ref{fig:slack_dist} shows an example: Even though the projection $P^{n}_{t+1}$ of the vehicle, which currently turns right, ends in another vehicle, it is not considered a violation.

\begin{figure}[tp]
\includegraphics[scale=0.36]{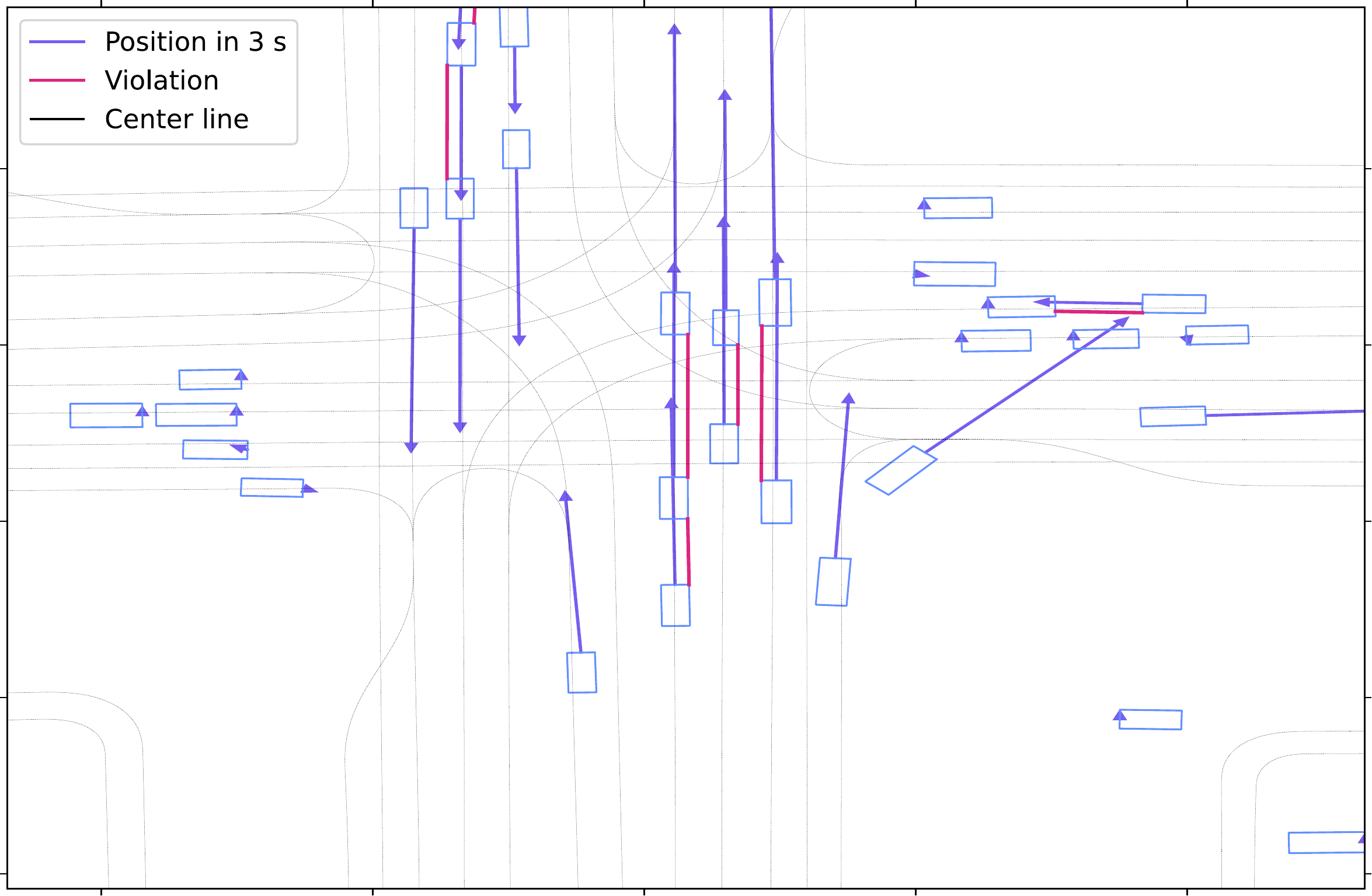}    
\caption{Visualization of the \textit{Safety Distance} rule analysis. Purple vectors refer to the vehicle's projected position three seconds into the future. If these vectors intersect with another vehicle, under certain conditions a pink line indicates a rule violation. The scene displays five drivers that are disobeying the \textit{Safety Distance} rule.}
\label{fig:slack_dist}
\end{figure}

\subsection{Rule Conformity for Speed Limits}
\label{subsection:speed_limit}
\begin{figure}[tp]
\includegraphics[scale=0.36]{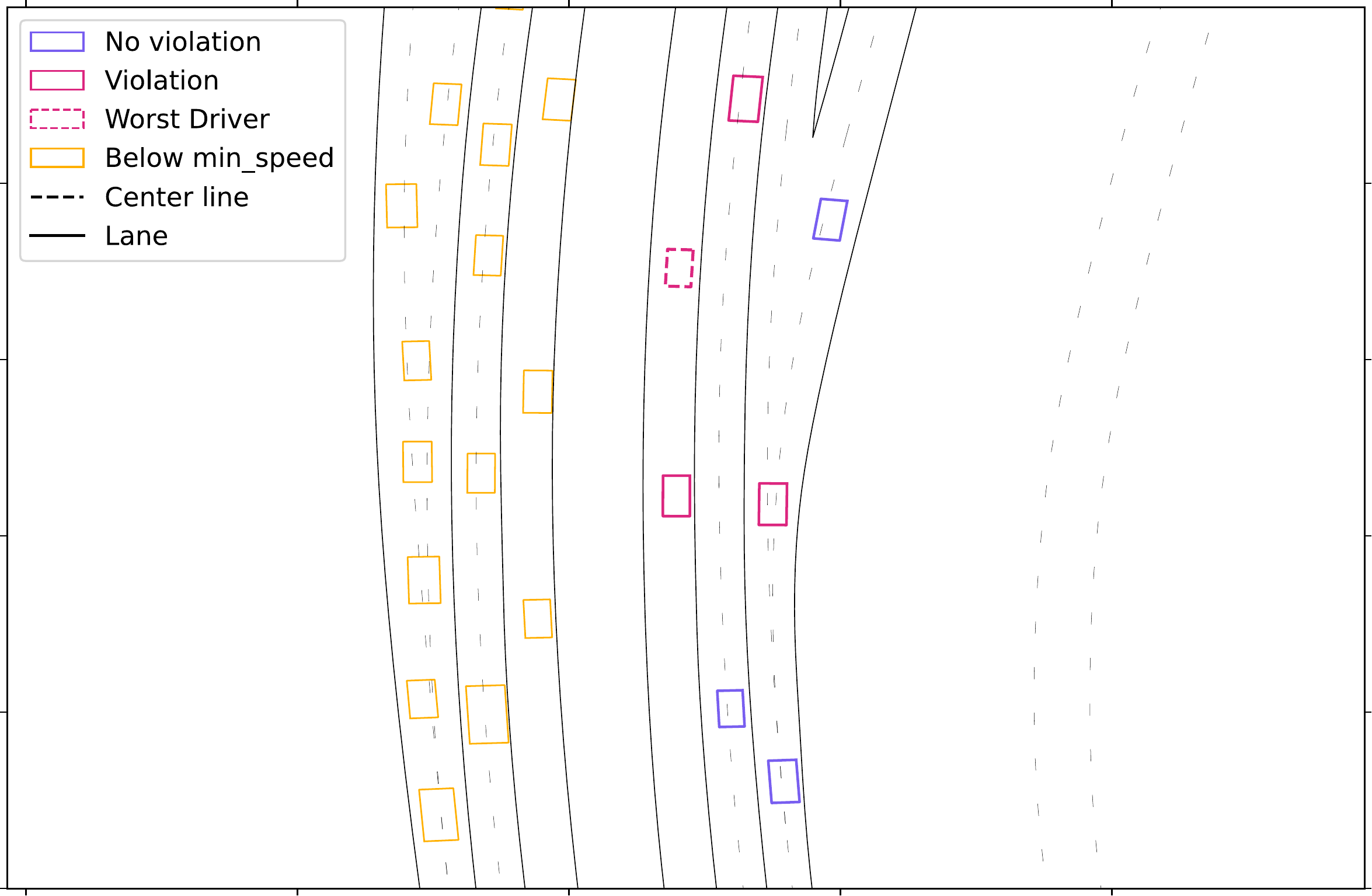}
\caption{Visualization of the \textit{Speed Limit} rule analysis. There are lanes in two directions, where the lanes on the left side are crowded. These vehicles are relatively slow and thus discarded from the quantification process and marked yellow. On the right side, vehicles that drive according to the speed limit appear violet, whereas cars above the limit are marked pink.}
\label{fig:speed_limit}
\end{figure}

To assign the currently effective \textit{Speed Limit} to individual vehicles, we took center lines into account, as these are assigned lane speed limit information in the Waymo map data. For every driver $n$, a lane $k$ is assigned based on the minimum euclidean distance. Since Waymo represents lanes as polylines, consisting of multiple points instead of defined vectors, it was necessary for each driver $n$ to calculate the euclidean distance to all points of the polylines.
We filtered out vehicles that exceed a distance threshold of $10~m$ to their assigned center line in order to avoid misclassifications. This large threshold is due to the fact that some lanes in the map data do not contain center lines. For instance, in figure \ref{fig:speed_limit}, all vehicles have the correct speed limit assigned even though center lines are missing. Furthermore, we omit vehicles below $80\%$ of the speed limit to exclude parked or congested vehicles. Afterwards, the \textit{Speed Limit} Conformity degree $RC^{speed}_{n,t}$ for driver $n$ at time $t$ is calculated: 
\begin{equation}
    RC^{speed}_{n,t} = min(1,\frac{limit_k}{v_{n,t}})
\end{equation}
Here, $limit_k$ denotes the speed limit of lane $k$ that corresponds to vehicle $n$ in state $t$. $v_n$ denotes the velocity of driver $n$. Accumulating these individual values, we retrieve an average rule conformity degree for every scenario in the dataset:
\begin{equation}
    RC^{speed}_{s} = \frac{1}{N}\sum_{n=0}^N\Bigl(\frac{1}{T}\sum_{t=0}^TRC^{speed}_{n,t}\Bigl)
\end{equation}

For the scene shown in Figure~\ref{fig:speed_limit}, $RC^{speed}_{s}\approx 0.9677$. It reveals the underlying filter process, that discards the crowded left lane, because of their slow velocity. On the right lane, the traffic is fluent and therefore utilized for the algorithm.
\section{Evaluation}
\label{sec:evaluation}
\newcommand\scale{0.385} 

In the previous sections we have introduced the Waymo Open Motion dataset, the two traffic rules \textit{Safety Distance} and \textit{Speed Limit} and our quantification methodology to derive the rule conformity per driver and scenario. In this chapter, we will extend the evaluation to the complete data set to obtain a general behavioral distribution. The process can be applied to any motion dataset to determine these behavior distributions for other, such as country-specific, domains.

After computing the average scores $RC^{dist}_{s}$, $RC^{speed}_{s}$ for each scenario $s$, we can compute the average over the total number of scenarios $S$ for both $RC^{dist}_{total}$ and $RC^{speed}_{total}$:


\begin{equation}
    RC_{total} = \frac{1}{S}\sum_{s=0}^SRC_s
\end{equation}

Since this value only represents the average over all scenarios per traffic rule, a more meaningful insight is the distribution of behaviors. To compute it, we performed the following steps per traffic rule:

\begin{enumerate}
    \item For each scenario, we calculated $RC_{n,t}$ for every driver $n$ at each frame $t$
    \item For each driver $n$ in a scenario $s$, we determined their average $RC_{s,n}=\frac{1}{T}\sum_{t=0}^TRC_{n,t}$ over all frames $t$
    \item We computed the distribution over all $RC_{s,n}$. Thus, every data point represents the averaged behavior of a single driver in a specific scenario. We visualized the data as histograms
\end{enumerate}

\subsection{Safety Distance}

\begin{figure}[tp]
    \centering
    \includegraphics[scale=\scale]{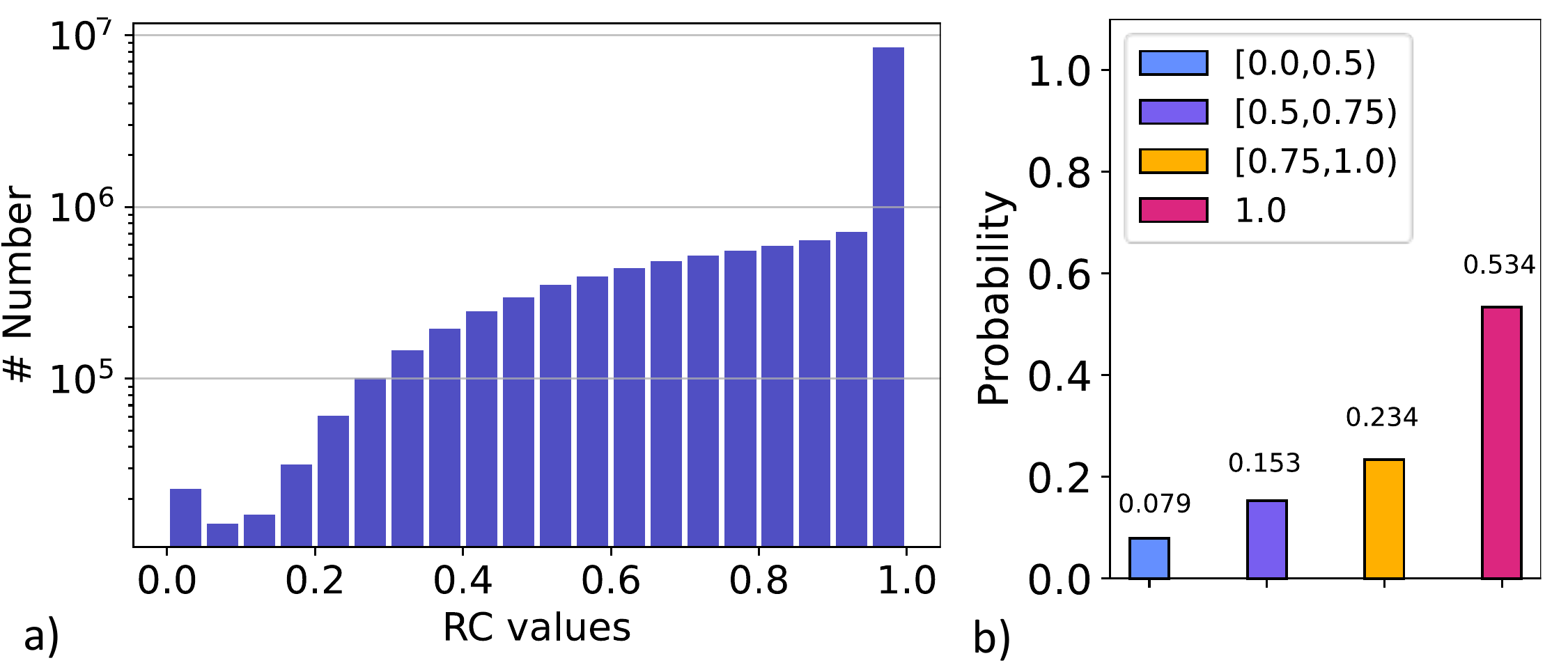}
    \caption{Distribution of the rule conformity in regard to \textit{Safety Distances}. In a), the absolute numbers of binned $RC$ values are shown on a logarithmic scale. In b), the relative amount of $RC$ values is summarized in four bins.}
    \label{fig:eval_dist}
\end{figure}


A clear picture emerges from the behavioral analysis, where we analyzed 14,317,443 data points. With $RC^{dist}_{total} = 0.867$, the data shows that many drivers do not strictly follow the \textit{Safety Distance} rule. This is in line with prior assumptions from own experiences. Figure~\ref{fig:eval_dist}~a) shows the behavior distribution regarding the compliance with the \textit{Safety Distance} rule in absolute numbers. We divided the $RC_{n,s}$ values into 20 bins and used a logarithmic scale to obtain a suitable histogram. As you can see in the histogram on the left, the numbers are slightly increasing again for the first bin. Our analysis shows that there are perception errors in the Waymo Open Motion data set. These scenes mostly show overlapping vehicle detections. Since we only used generic rules for our pre-processing instead of correcting dataset-specific errors, these also show up in the final distribution.

The rightmost bin at $1.0$ displays the number of drivers that follow \textit{Safety Distance} rule strictly, whereas the other bins visualize the magnitude of rule bending. On the other side, there are hardly any deviations below a value of $0.4$ . 


In order to analyze the relative amount of local rule conformity scores, we generated a percentage plot as shown in Figure~\ref{fig:eval_dist}~b). It shows with the rightmost bar that over $50~\%$ of all drivers adhere to the rule strictly, while drivers bend the distance rule to some extent in all other cases. However, strong violations are rather seldom, since the interval $[0.75,1.0]$, or the orange and pink bar respectively, add up to almost 80\%. To conclude, the distribution shows that most drivers follow the rule strictly or only with minor deviations.


\subsection{Speed Limit}

\begin{figure}[tp]
    \centering
    \includegraphics[scale=\scale]{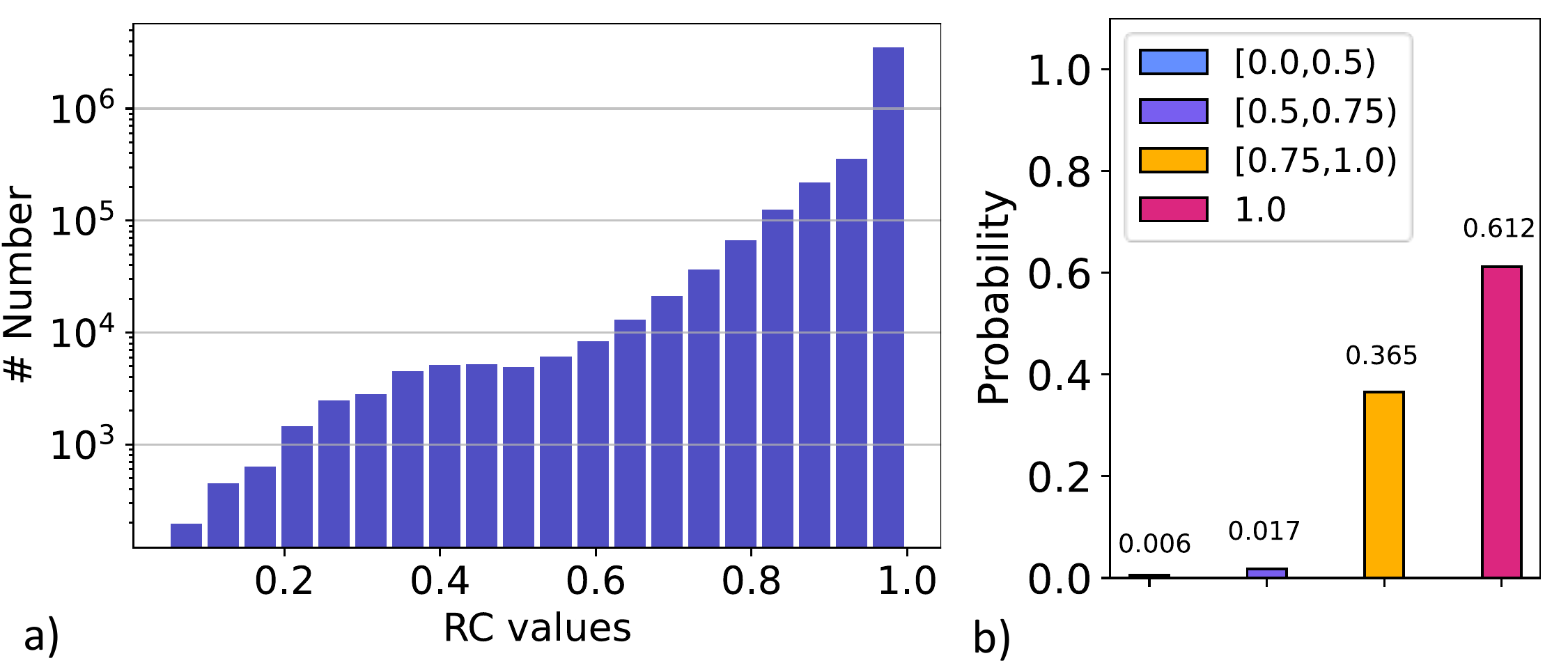}
    \caption{Distribution of the rule conformity in regard to \textit{Speed Limits}. In a), the absolute numbers of binned $RC$ values are shown on a logarithmic scale. In b), the relative amount of $RC$ values is summarized into four bins.}
    \label{fig:eval_speed}
\end{figure}

Unlike \textit{Safety Distance}, \textit{Speed Limit} is a highly controlled traffic rule. 
With a $RC^{speed}_{total} = 0.967$ the results of Figure~\ref{fig:eval_speed}~a) reveal that the global degree of rule compliance is relatively close to $1.0$. Here, we analyzed 4,400,350 data points. The significantly lower magnitude of the bin height compared to Figure \ref{fig:eval_dist} is a result of our extensive pre-processing. As mentioned in \ref{subsection:speed_limit} we filtered out most of the drivers, that are either waiting on red lights or are stuck in traffic jam, which reduces the number of tracked vehicles.
In contrast to the \textit{Safety Distance} rule, we encountered almost only $RC$ values above $0.75$ among the dataset, as shown in Figure~\ref{fig:eval_speed}~b). Values below this were rather rare. Thus, drivers usually drive no more than $25\%$ above the speed limit. Nevertheless, the relative plot in Figure~\ref{fig:eval_speed}~b) shows, that in more than $60~\%$ of the time, the rule is strictly observed. Contrary to the results of the \textit{Speed Limit} rule, we did not expect this behavior. The reasons for this could be a general speed limit and significantly higher penalties for exceeding it. This indicates a domain gap compared to the behavior on German roads. To conclude, the deviations in regard to the \textit{Speed Limit} rule are significantly lower than for the \textit{Safety Distance} rule. Especially severe violations are extremely rare among the dataset.

\section{Conclusion}
\label{sec:conclusion}

Traffic rules are static constructs that aim to guide road users safely through traffic. In doing so, human road users usually do not adhere strictly to these rules, but with certain deviations. In this work, we developed a framework to quantify human behavior in respect to given traffic rules for motion datasets. We demonstrated our approach based on the Waymo Open Motion Dataset. We analyzed the behavior regarding two rules, namely \textit{Safety Distance} and \textit{Speed Limits}, for which we extracted behavior distributions and averages for scenarios and the whole dataset. In addition, we took some pre-processing steps to keep the data as meaningful as possible and to remove noise. All our code including extraction, pre-processing and visualization is publicly available at \url{https://url.fzi.de/iv2022}. 

We found that there is significantly more deviation in respect to the \textit{Safety Distance} rule than for the \textit{Speed Limit} rule. In addition, the magnitude of violations is larger in the \textit{Safety Distance} rule, as there are more strong violations. However, the \textit{Speed Limit} rule is also frequently deviated from, thus both rules are not strictly followed.

These results made us confident that additional knowledge on human behavior instead of only relying on given, static traffic rules is necessary for autonomous vehicles to perform more flexible and in a more human-like style. When designing a system that is limited by rules, it is of utmost importance to also analyze the degree of such restrictions. Such investigations reveal the true importance of a rule, which could differ from biased developer assumptions.

With parallelized algorithms, we want to analyze the complete data set instead of the 1 Hz samples in the future. Additionally, the analysis in respect to further rules is of interest, especially if a large and diverse dataset recorded in Germany might be available in the future. Such extensions should closely follow the approach presented in this paper. For this purpose, a formal definition of the rules should be made in each case and these should be made quantifiable. The extraction must be performed on as large dataset as possible, e.g., from Waymo, in order to obtain a meaningful statement about the underlying rule conformity of drivers. New rules can add a significant amount of implementation overhead and significantly increase runtime. Regarding applications of this framework, we envision many opportunities. The detection of interesting scenes and scenario based on the scenario-based degrees of rule conformity is one example. Also, based on the global distribution, rewards in Reinforcement Learning can be shaped to improve human-like behavior.









\section{Acknowledgment}
\label{sec:acknowledgment}

This work results partly from the project KI WISSEN (19A20020L), funded by the German Federal Ministry for Economic Affairs and Climate Action (BMWK).

{\small
\bibliographystyle{IEEEtran}
\bibliography{references}
}

\end{document}